\documentclass[]{lab}

\usepackage[utf8]{inputenc}
\usepackage[T1]{fontenc}
\usepackage{geometry}
\usepackage{amsmath,amssymb}  %
\usepackage{charter}

\usepackage[toc,page,header]{appendix}

\usepackage{minitoc}
\usepackage{cleveref} 
\usepackage{subcaption}
\usepackage{booktabs}
\usepackage{graphicx}
\usepackage{pgfplots}
\usepackage{pgfplotstable}
\usepackage{xcolor}
\usepackage{CJKutf8}
\usetikzlibrary{patterns}
\usepackage{float}
\usepackage{placeins}
\usepackage{caption}
\usepackage{bm}

\usepackage{soul} %
\usepackage{algorithm}      %
\usepackage{algpseudocode}  %
\usepackage{parskip}

\definecolor{cvprblue}{rgb}{0.21,0.49,0.74}
\definecolor{fallbackgreen}{rgb}{130, 180, 102}
\definecolor{stopred}{rgb}{251, 225, 224}

\ifdefined\final
\usepackage[disable]{todonotes}
\else
\usepackage[textsize=tiny]{todonotes}
\fi

\newcommand{\ours}{\texttt{Motubrain}\xspace}

\usepackage{natbib}
\usepackage{latexsym}

\usepackage{url}
\usepackage{amssymb}
\usepackage[utf8]{inputenc}
\usepackage{microtype}
\usepackage{booktabs}
\usepackage{pifont} 
\usepackage{multirow}
\usepackage{makecell}
\usepackage{paralist}
\usepackage{xspace}
\usepackage{color}
\usepackage{xcolor}
\usepackage{colortbl}
\usepackage{adjustbox}
\usepackage{hyperref} 
\usepackage[edges]{forest}
\usepackage{tikz} 
\usepackage{caption}
\usepackage{amsfonts}

\hypersetup{
    colorlinks,
    linkcolor={blue!80!black},
    citecolor={blue!80!black},
}
\tikzset{
    root/.style =             {align=center, text width=1cm, rounded corners=3pt, line width=0.3mm, fill=gray!10, draw=gray!80, font=\small},
    demographic/.style =         {align=center, text width=1.8cm, rounded corners=3pt, line width=0.3mm, fill=blue!10, draw=blue!80, font=\footnotesize},
    demographic_work/.style =    {align=center, text width=10cm, rounded corners=3pt, line width=0.3mm, fill=blue!10, draw=blue!0, font=\footnotesize},
    character/.style =         {align=center, text width=1.8cm, rounded corners=3pt, line width=0.3mm, fill=red!10, draw=red!80, font=\footnotesize},
    character_work/.style =    {align=center, text width=10cm, rounded corners=3pt, line width=0.3mm, fill=red!10, draw=red!0, font=\footnotesize},
    personalization/.style =           {align=center, text width=1.8cm, rounded corners=3pt, line width=0.3mm, fill=cyan!10, draw=cyan!80, font=\footnotesize},
    personalization_work/.style =      {align=center, text width=10cm, rounded corners=3pt, line width=0.3mm, fill=cyan!10, draw=cyan!0, font=\footnotesize},
    risk/.style =         {align=center, text width=1.8cm, rounded corners=3pt, line width=0.3mm, fill=orange!10, draw=orange!80, font=\footnotesize},
    risk_work/.style =    {align=center, text width=10cm, rounded corners=3pt, line width=0.3mm, fill=orange!10, draw=orange!0, font=\footnotesize},
}

\usepackage{CJK}

\newtcolorbox{promptbox}[1][]{
  enhanced,
  breakable,
  colback=promptboxlightgray,
  colframe=promptboxblue!30,
  arc=8pt,
  boxrule=0.5pt,
  left=12pt,
  right=12pt,
  top=8pt,
  bottom=8pt,
  fonttitle=\bfseries,
  fontupper=\linespread{1.2}\selectfont,
  title=#1
}

\title{Motubrain: An Advanced World Action Model\\ for Robot Control}

\author{Motubrain Team \\Project Page: \url{https://www.genspi.com/en/motubrain}\\GitHub: \url{https://github.com/shengshu-ai/Motubrain}}

\begin{document}

\maketitle


\section{Introduction}

Recent progress in embodied foundation models has been driven by Vision-Language-Action (VLA) policies~\cite{intelligence2025pi05visionlanguageactionmodelopenworld,black2026pi0visionlanguageactionflowmodel, liu2025rdt1bdiffusionfoundationmodel,kim2024openvlaopensourcevisionlanguageactionmodel, zheng2025xvlasoftpromptedtransformerscalable, bu2025univlalearningacttaskcentric, octomodelteam2024octoopensourcegeneralistrobot, brohan2023rt2visionlanguageactionmodelstransfer},  which map visual observations and language instructions in pretrained Vision-Language Models (VLMs)~\cite{beyer2024paligemma,bai2025qwen3} directly to robot actions. Inheriting rich semantic priors, VLAs generalize well across diverse objects and language instructions, achieving strong performance on a wide range of robotic tasks. However, since they are primarily pretrained on static image–text data, they often neglect the perception and prediction of fine-grained world dynamics that are essential for precise robotic control, leading to a superficial imitation of behaviors rather than temporal understanding of the physics of the world.

With the rise of video generation models~\cite{wan2025,bao2024viduhighlyconsistentdynamic,zheng2024opensorademocratizingefficientvideo,seedance2026seedance20advancingvideo,kong2025hunyuanvideosystematicframeworklarge}, a growing body of research has begun exploring how to adapt these models for world modeling~\cite{gao2025adaworldlearningadaptableworld,he2025pretrainedvideogenerativemodels,nvidia2025cosmosworldfoundationmodel, bruce2024geniegenerativeinteractiveenvironments,he2026matrixgame20opensourcerealtime,guo2025ctrl}. A world model aims to predict how the environment evolves in response to actions—a capability that aligns directly with what a video generation model does when conditioned on past observations and actions, namely forecasting future visual states. Conceptually, video generation models are naturally suited to this task, as video generation models are pretrained on vast and diverse web video data, which equips them with rich spatiotemporal priors regarding object permanence, physical dynamics, and human-object interactions. This enables them to generalize more effectively and reason about novel scenarios where in-domain data are scarce. As a result, leveraging video generation models as a foundation for world modeling has emerged as a promising paradigm for scalable robot policy learning.

Following this insight, early attempts largely followed a two-stage paradigm: Video Generation Model (VGM) plus Inverse Dynamics Model (IDM)~\cite{feng2025vidar,hu2024video,zhou2024robodreamerlearningcompositionalworld, pai2025mimicvideovideoactionmodelsgeneralizable, du2023learninguniversalpoliciestextguided}. In this framework, a video diffusion model pretrained on large-scale web data first predicts future visual trajectories from current observations and language instructions. Subsequently, an inverse dynamics model infers actions from the generated future frames. While this paradigm successfully leverages rich spatiotemporal priors from video data to achieve broad generalization, it suffers from a critical drawback in that errors in video prediction accumulate over time, which leads to compromised action accuracy and downstream policy performance.

To mitigate this issue, a subsequent line of research has explored World Action Models (WAMs) that unify visual dynamics and action prediction within a single, jointly optimized objective~\cite{bi2025motusunifiedlatentaction,yuan2026fast,ye2026world,kim2026cosmos,liao2025genie,zhu2025uwm,li2026causal,lyu2026lda1bscalinglatentdynamics}. Unlike the VGM+IDM pipeline, which decouples forecasting from action inference, WAMs simultaneously predict future visual states and actions in an aligned manner. This integration offers two key advantages: (1) it avoids the cascading errors inherent in sequential video prediction, and (2) it overcomes the fragmented functionality of conventional embodied systems, where semantic understanding, dynamics modeling, and action generation are typically learned from disparate supervision sources.

In our view, the core source of intelligence in such a unified model is its ability to absorb large-scale heterogeneous multimodal data under one unified training recipe. In principle, this includes pure video data without action annotations, robot data with aligned video-language-action trajectories across different embodiments, and even task-agnostic interaction data with partially missing modalities~\cite{tan2025anyposautomatedtaskagnosticactions}. By contrast, VLA learning primarily relies on robot task trajectories with aligned observation-language-action supervision, and adaptation to the target robot is primarily coupled to embodiment-specific action data~\cite{kim2024openvlaopensourcevisionlanguageactionmodel,black2026pi0visionlanguageactionflowmodel,openx2023embodiment}.

Motus~\cite{bi2025motusunifiedlatentaction} was an early step in this direction. It established a unified world-action formulation in which video and action are modeled in a shared generative framework, so that policy modeling, world modeling, video generation, inverse dynamics, and joint video-action prediction become different inference modes of the same model. It also showed that, with UniDiffuser-style continuous multimodal modeling and a Mixture-of-Transformers design, a world action model can absorb heterogeneous multimodal data rather than being restricted to embodiment-specific task trajectories.

Building on this foundation, we present \ours. Like Motus, \ours adopts UniDiffuser~\cite{bao2023transformerfitsdistributionsmultimodal} to jointly model and schedule the two continuous modalities, namely video and action, and uses a three-stream Mixture-of-Transformers architecture to integrate video generation, action modeling, and language conditioning within one system. This unified formulation again supports inference over five distributions with the same model: vision-language-action policy modeling, world modeling, video generation, inverse dynamics, and joint video-action prediction. More importantly, it preserves the key advantage of unified world-action modeling: the model can learn from a much broader family of multimodal data, including video-only data without action labels, interaction data without explicit task language, and heterogeneous robot trajectories collected from different embodiments.

\ours further extends this paradigm in several practically important directions. It introduces a unified multiview representation that supports an arbitrary number of camera views under different camera layouts, rather than depending on a fixed visual input format. It uses an independent text stream to more tightly couple high-level semantics with low-level control, making instruction following an explicit part of action generation and improving semantic understanding. It adopts a unified action representation across embodiments, enabling the model to capture transferable control regularities rather than overfitting to the action format of a single robot. Beyond architecture and pre-training, \ours also involves a post-training and deployment recipe tailored to long-horizon real-world control: autoregressive diffusion rollout enables temporally extended execution, V2A-style asymmetric dependency enables action-only inference without explicitly generating future video, and real-time chunked closed-loop execution reduces boundary discontinuities during asynchronous control. Finally, we develop a systems-oriented inference stack including denoising-step reduction, compilation with CUDA-graph-friendly execution, FP8 quantization, and DiT caching, which together deliver more than $50\times$ end-to-end speedup over the naive baseline and make large world action models practical for real-time robotic deployment.

These design choices lead to strong empirical performance across both action-centric and world-modeling evaluations. On RoboTwin 2.0, \ours achieves average success rates of $95.8\%$ and $96.1\%$ under the clean and randomized settings, respectively, with the randomized score exceeding $95\%$. On WorldArena, \ours attains the strongest reported EWMScore in our comparison, indicating that the same model not only executes actions effectively but also predicts future world dynamics accurately. Beyond standardized benchmarks, \ours can be adapted to new humanoid embodiments with only 50--100 same-embodiment trajectories, while retaining the ability to solve long-horizon and dexterous manipulation tasks without relying on an additional VLM planner, a dual-system decomposition, external memory, or retry-specific data. Together, these results suggest that unified world action models can simultaneously scale in generality, controllability, and real-world deployability.


\begin{figure}[t]
  \centering
  \includegraphics[width=\linewidth]{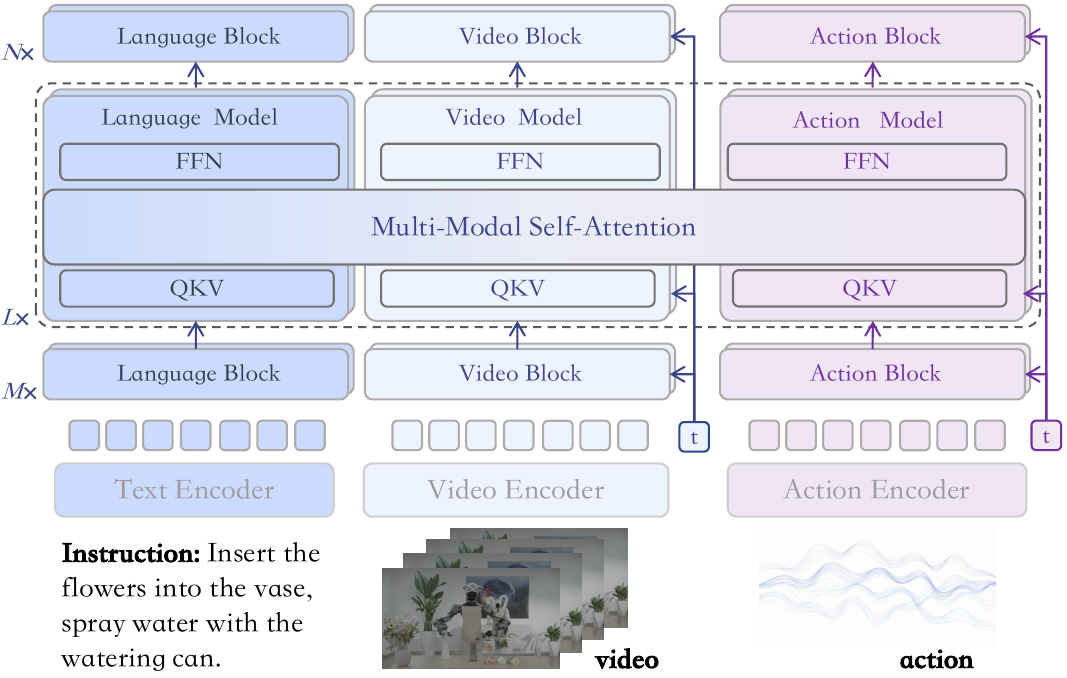}
  \caption{Overview of \ours's architecture. \ours builds on a unified video-action backbone and adopts a three-stream Mixture-of-Transformers architecture with text, video, and action streams. It further uses an H-bridge attention design to balance cross-modal interaction and efficiency, while supporting flexible multiview inputs through view-dependent 3D RoPE offsets.}

\end{figure}

\section{Method}

In this section, we present the overall methodology of \ours from four aspects: model architecture, pre-training, post-training, and inference. We first introduce the core architectural design and the main acceleration techniques integrated into \ours in Section~\ref{sec:model_arch}. We then describe the pretrained foundation model used in our system in Section~\ref{sec:pretraining}. After that, we present the post-training pipeline, including sparse adaptation and step distillation, in Section~\ref{sec:posttraining}. Finally, we describe the inference-time acceleration strategies deployed in \ours in Section~\ref{sec:inference}.

\subsection{Model Architecture} \label{sec:model_arch}

Motubrain first adopts UniDiffuser~\cite{bao2023transformerfitsdistributionsmultimodal} to jointly model and schedule the two continuous modalities, i.e., video and action, so that all interaction patterns between them are captured in a unified generative framework. As a result, a single training procedure supports inference over five distributions: vision-language-action policy modeling, world modeling, video generation, inverse dynamics, and joint video-action prediction, which are formulated in Table~\ref{tab:prediction_formulations}.

\begin{table}[t]
\centering
\caption{Different prediction modes (non-autoregressive mode).}
\label{tab:prediction_formulations}
\begin{tabular}{c|c}
\hline
Model & Prediction Objective \\
\hline
VLA & $p(\bm{a}_{t+1:t+k} \mid \bm{o}_t, \ell)$ \\
WM & $p(\bm{o}_{t+1:t+k} \mid \bm{o}_t, \bm{a}_{t+1:t+k})$ \\
IDM & $p(\bm{a}_{t+1:t+k} \mid \bm{o}_{t:t+k})$ \\
VGM & $p(\bm{o}_{t+1:t+k} \mid \bm{o}_t, \ell)$ \\
Joint Video-Action Prediction & $p(\bm{o}_{t+1:t+k}, \bm{a}_{t+1:t+k} \mid \bm{o}_t, \ell)$ \\
\hline
\end{tabular}
\end{table}

On top of this unified video-action backbone, \ours introduces a three-stream Mixture-of-Transformers (MoT) architecture with a text stream, a video stream, and an action stream. The text stream serves as a conditioning branch: its hidden states participate in Transformer attention, but no output head is applied to text tokens. Maintaining a dedicated text stream also improves semantic understanding and instruction-following capability. The video and action streams are both trained with flow matching, predicting the velocity fields of video latents and action tokens, respectively.

The inputs consist of text tokens, condition-image latents encoded by a Vidu VAE, noisy future video latents, and noisy action tokens. The condition image is represented as the first video latent frame and is teacher-forced in the video stream, while the remaining video latents and action tokens are denoised by the video and action streams. Cross-modal interaction is implemented through joint attention over video, action, and text tokens.

Instead of using full video-action joint attention in all layers, we adopt an H-bridge design following HBridge~\cite{wang2025hbridge}. Concretely, full V-A joint attention is applied only in the middle 50\% of Transformer layers, while the bottom and top 25\% layers use
decoupled attention. In the decoupled layers, video tokens and action tokens are processed independently and do not perform joint-attention. This design reduces the cost of dense cross-modal attention, improves efficiency, preserves modality-specific representations in shallow and deep layers, and avoids injecting excessive modality-irrelevant information into every layer, while still allowing semantic alignment and policy grounding in intermediate layers.

For multiview inputs, each view is encoded independently by the Vidu VAE and concatenated at the token level. Leveraging 3D RoPE, we introduce view-dependent offsets only along the spatial dimensions while keeping the temporal dimension unchanged. This effectively maps different views to distinct regions in a shared spatial positional space, enabling seamless support for an arbitrary number of camera views without modifying the backbone architecture.


\subsection{Pre-training} \label{sec:pretraining}

Our pre-training data follows a four-level data pyramid~\cite{bi2025motusunifiedlatentaction}, organized from broad visual diversity to embodiment-specific control signals: Internet videos, ego-centric videos, heterogeneous-embodiment data, and specific-embodiment data. The main motivation is to maximize data efficiency. The bottom level uses large-scale Internet videos to train the video generation model Vidu~\cite{bao2024viduhighlyconsistentdynamic}, which serves as the foundation model of \ours. The second level introduces ego-centric videos, which provide first-person interaction patterns and hand-object dynamics that are closer to embodied manipulation. The third level uses heterogeneous-embodiment data collected from different robot platforms, tasks, and scenes. In our setting, we only use dual-arm robot data at this level. The top level consists of specific-embodiment data collected on the target robot configuration, which further aligns the model with the final action space, kinematics, camera setup, and deployment distribution. This hierarchy reflects a central design principle of \ours: the model should learn from heterogeneous multimodal data wherever available, rather than restricting supervision to a single narrow data format.

Starting from the pretrained Vidu weights~\cite{bao2024viduhighlyconsistentdynamic}, we perform two-stage pre-training corresponding to the second and third levels of the data pyramid. In the first stage, we train only the video branch on ego-centric and heterogeneous-embodiment data, while keeping the randomly initialized action branch unchanged. Accordingly, the optimization objective in this stage contains only the video loss. This stage is designed to adapt the Internet-scale video prior to embodied manipulation and to obtain a video world model that can understand and predict bimanual interaction dynamics. To improve robustness to imperfect visual conditioning, we follow the noisy-conditioning strategy of LingBot-VA~\cite{li2026causal} throughout training, including stage 1, stage 2, and non-autoregressive post-training, but not the autoregressive policy setting. In addition, for multiview data, we randomly drop auxiliary views with probability $0.1$ during pre-training, so that the model can better adapt to varying numbers of camera views and imperfect visual observations. Concretely, with probability $p=0.5$, we perturb the conditioned-frame latent as
\begin{equation}
\tilde{z}_{0} = s_{\mathrm{aug}} z_{0} + (1-s_{\mathrm{aug}}) \epsilon, \qquad s_{\mathrm{aug}} \sim \mathcal{U}[0.3,0.7], \quad \epsilon \sim \mathcal{N}(0,I),
\end{equation}
and leave it unchanged otherwise. In the second stage, we initialize from the first-stage checkpoint and train only the action branch on heterogeneous-embodiment data, while freezing the video branch. At this stage, we use a unified action representation across embodiments.

Concretely, let the absolute end-effector chunk be $E^{\mathrm{abs}}=\{e^{\mathrm{abs}}_1,\ldots,e^{\mathrm{abs}}_n\}$ and let $s$ denote the end-effector state of the conditioned frame. We define the corresponding relative chunk as $E^{\mathrm{rel}}=\{e^{\mathrm{rel}}_1,\ldots,e^{\mathrm{rel}}_n\}$, where each action is represented as
\begin{equation}
e^{\mathrm{rel}}_i = e^{\mathrm{abs}}_i \ominus s.
\end{equation}
Here $\ominus$ denotes a component-wise pose difference: the position is computed by direct subtraction, the rotation is computed by composition with the inverse reference rotation, and the gripper state is kept unchanged. Writing $e=(p,R,g)$ with position $p$, rotation $R$, and gripper state $g$, we have
\begin{equation}
e^{\mathrm{abs}}_i=(p_i,R_i,g_i), \qquad s=(p_s,R_s,g_s),
\end{equation}
and
\begin{equation}
e^{\mathrm{rel}}_i=(p_i-p_s,\; R_s^{-1}R_i,\; g_i).
\end{equation}
The raw input pose is provided in quaternion format, while the training target uses a $6$D rotation representation. Each end-effector action therefore has dimension $10$, consisting of position, rotation, and gripper state. We only normalize the gripper dimension to $[-1,1]$ and keep the remaining dimensions in their original physical scales. Using relative end-effector coordinates with respect to the conditioned frame improves compatibility across different robot embodiments and initial poses, making the action space more consistent under heterogeneous pre-training and allowing the target embodiment to be adapted with less embodiment-specific data. Although only the action branch is updated, stage 2 still optimizes both video and action objectives under the unified formulation. We use separate SNR-based timestep sampling for the two modalities, with $\mathrm{timeshift}=6$ for video and $\mathrm{timeshift}=1$ for action. The overall stage-2 objective is the weighted sum of the video and action losses,
\begin{equation}
\mathcal{L} = \lambda_v \, \mathcal{L}_v + \lambda_a \, \mathcal{L}_a,
\end{equation}
where
\begin{equation}
\mathcal{L}_v = \mathrm{MSE}(v_{\mathrm{out}}, v_{\mathrm{target}}), \qquad
\mathcal{L}_a = \mathrm{MSE}(a_{\mathrm{out}}, a_{\mathrm{target}}).
\end{equation}
This design encourages alignment between the action and video modalities within the unified model: action learning benefits from the visual dynamics encoded by the video branch, while the action-conditioned interaction also improves video prediction by injecting control-relevant information into the shared modeling process.

\begin{figure}[ht]
  \centering
  \includegraphics[width=\linewidth]{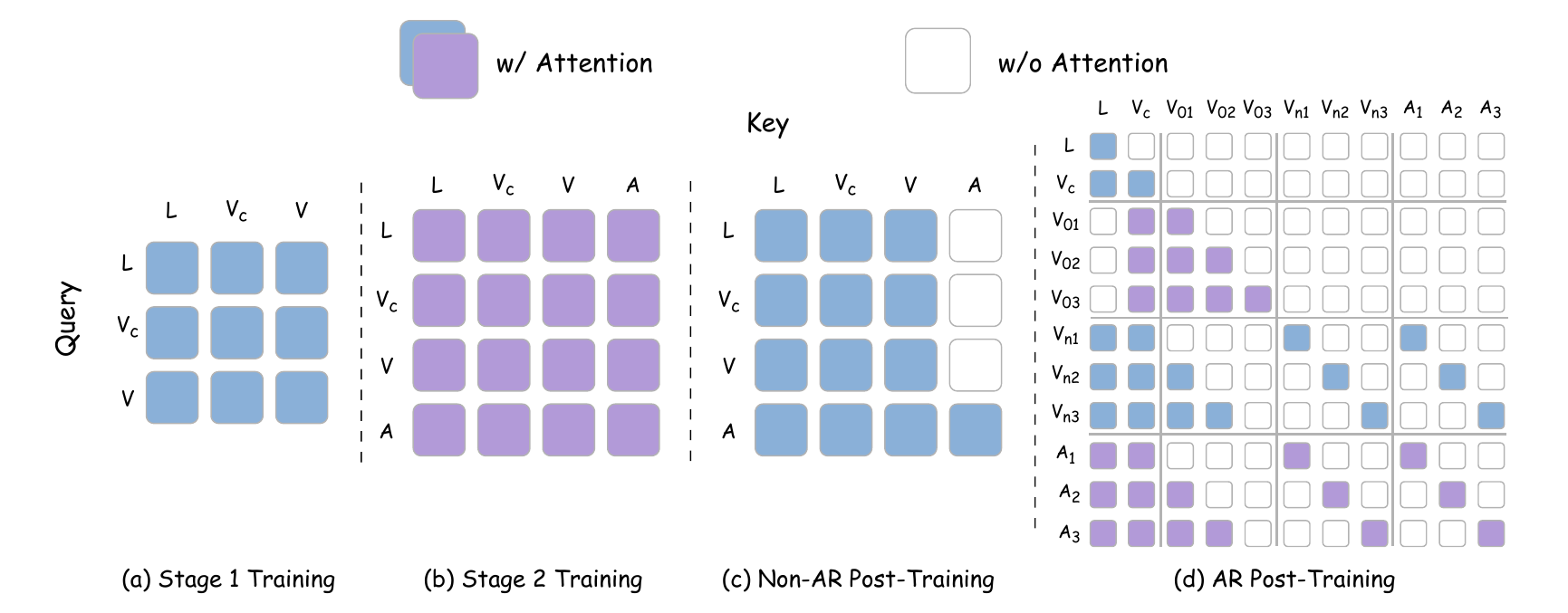}
  \caption{Attention masks used in all training stages and post-training modes. (a) Stage-1 pre-training updates only the video branch. (b) Stage-2 pre-training uses full joint attention over language, conditioned video, video, and action tokens. (c) Non-autoregressive post-training disables action-to-video attention. (d) Autoregressive post-training applies a causal mask over temporally ordered video and action tokens. $V_0$ denotes clean video tokens while $V_n$ denotes noisy video tokens to be denoised.}
  \label{fig:attn_map}
\end{figure}

\subsection{Post-training} \label{sec:posttraining}

While pre-training aims to build a unified world action model from broad and heterogeneous data, post-training focuses on adapting the model to the target embodiment. We start from the checkpoint obtained after the second stage of pre-training and finetune it on specific-embodiment data. In this stage, we consider two post-training settings, namely a non-autoregressive (Non-AR) setting and an autoregressive (AR) setting, and train separately under each setting. The attention masks used in stage 1, stage 2, Non-AR, and AR are summarized in Figure~\ref{fig:attn_map}. The two post-training settings mainly differ in sequence factorization and attention masking.

Under the Non-AR setting, the model denoises all video and action tokens within the full observation window in a single forward pass, where $o_t$ denotes the observation at step $t$, $z_t = \mathcal{E}(o_t)$ is its VAE latent, the VAE temporally compresses every $\tau$ consecutive video frames into a single latent frame, and the action stream operates at $f_{va}$ actions per raw video frame. Each latent frame therefore corresponds to $S_a = f_{va}\tau$ action tokens. Consequently, predicting $K$ future video latents amounts to generating $K S_a$ actions, enabling high-frequency control while keeping the visual stream compact. As noted in Section~\ref{sec:pretraining}, we continue to apply the same LingBot-VA-style noisy-conditioning augmentation to the conditioned-frame latent in Non-AR post-training. This improves robustness to noisy conditioning and helps the model recover from partially corrupted observations.

For long horizon tasks, we adopt an AR formulation based on chunk-level factorization, as illustrated in Figure~\ref{fig:attn_map}(d). We partition each episode into non-overlapping chunks and process all chunks in parallel during training with a block-causal attention mask. The input sequence contains language tokens, conditioned-image tokens, clean video tokens from teacher-forced past observations, and the noisy video and noisy action tokens of the current target chunk. Unlike Non-AR post-training, the AR setting does not apply the noisy-conditioning augmentation to the conditioned frame. For chunk $k$, tokens can attend only to clean visual context from preceding chunks, but never to future chunks. Importantly, our AR model does not involve clean action tokens, because doing so would break the unified relative-EEF action representation. At deployment, the model rolls out sequentially, using the newly observed frame as clean context for the next chunk.

To better align post-training with deployment, we adopt V2A-style attention in both operating modes. Under this attention pattern, action tokens attend to video and language tokens, while video tokens never attend to action tokens. Combined with the UniDiffuser formulation, where video and action are sampled with independent timesteps, this asymmetric dependency makes it possible to use an action-only suffix during inference: after a short joint denoising prefix, the video stream can be frozen and only the action stream continues to update while attending to the cached visual-language context. This design is motivated by efficiency rather than a change in training objective, and in practice it enables substantial inference acceleration without degrading task success rate.

\subsection{Inference} \label{sec:inference}

\subsubsection{Inference Optimization}

A practical concern with world-action models is inference latency: jointly denoising a high-dimensional video latent and an action latent over many diffusion steps couples two of the slowest computations in modern generative modeling, and naively deployed WAMs typically run far below the control rates required for high frequency manipulation. We adopt and extend a stack of optimizations to tackle this problem. In our deployment setting, inference is executed remotely on cloud GPUs rather than on-board the robot, and the optimized system reaches a real-time inference frequency of 11\,Hz. Crucially, we verify on the RoboTwin2.0 and real-world tasks that this speedup is essentially lossless: average task success rates fluctuate within sub-percent margins across the optimized and unoptimized configurations, indicating that the gains come from removing genuinely redundant computation rather than from sacrificing model fidelity. We detail each component below.

\paragraph{Noise sampling.}
We use separate SNR-based timestep sampling for the two modalities during training. Specifically, we set $\mathrm{timeshift}=6$ for video and $\mathrm{timeshift}=1$ for action. This makes video timesteps more likely to be sampled from noisier regions, while action timesteps are sampled more uniformly. As a result, the model becomes more robust at predicting accurate actions under noisy visual conditions. This improves both robustness and convergence, allowing the number of inference steps to be reduced from 50 to 30 without performance degradation.

\paragraph{Compiling.}
We optimize the inference graph and fuse the operators with \texttt{torch.compile} to reduce the overhead of repeated denoising. Since the model is rewritten as a single-GPU, pure-PyTorch inference model, the main Transformer computation can be compiled directly at inference time. This optimization primarily improves execution efficiency for the repeated DiT forwards used during sampling.

\paragraph{DiT Cache.}
We adopt a DreamZero-style DiT caching strategy~\cite{ye2026world} to exploit temporal redundancy across denoising steps. Let $v_t$ denote the velocity prediction at denoising step $t$. We measure the similarity between two consecutive predictions by
\begin{equation}
s_t = \frac{\langle v_t, v_{t-1} \rangle}{\|v_t\|_2 \, \|v_{t-1}\|_2}.
\end{equation}
When $s_t > \gamma$, where $\gamma$ is a predefined threshold, we skip a small number of subsequent DiT evaluations and approximate the skipped predictions from recent history, i.e.,
\begin{equation}
\hat{v}_{t+j} \approx v_t, \qquad j=1,\ldots,k,
\end{equation}
where $k$ is the cache length. The cache can be applied to either the video velocity or, in action-only mode, the action velocity, and is reset for each inference call or chunk.

\paragraph{FP8 Quantization.}
We further reduce DiT inference cost with FP8 quantization. The implementation replaces eligible \texttt{nn.Linear} layers with FP8 linear layers. We store the weights in \texttt{float8\_e4m3fn} with a per-tensor scale, dynamically quantize activations to FP8 at runtime, and compute matrix multiplications through \texttt{torch.\_scaled\_mm}, returning outputs in the original compute dtype. Layers whose input or output dimensions are not divisible by 16 are skipped to satisfy the kernel alignment requirement. This optimization mainly targets the large linear projections in attention and MLP blocks, reducing memory bandwidth and GEMM cost on FP8-capable GPUs. Since quantization is applied after checkpoint loading and before compilation, the compiled graph traces the quantized linear operators directly.

\paragraph{V2A-Style Inference.}
For models trained with V2A attention, the video stream is prevented from attending to action tokens, while the action stream can still attend to visual and language context. This asymmetric dependency enables an efficient inference schedule. Instead of denoising video and action jointly for all sampling steps, we use a short joint denoising prefix followed by an action-only suffix:
\begin{equation}
(z_v^{(t+1)}, z_a^{(t+1)}) =
\begin{cases}
\Phi_{\mathrm{joint}}\!\left(z_v^{(t)}, z_a^{(t)}\right), & t < N, \\
\left(z_v^{(N)},\, \Phi_{\mathrm{act}}\!\left(z_a^{(t)}; z_v^{(N)}\right)\right), & t \ge N,
\end{cases}
\end{equation}
where $z_v^{(t)}$ and $z_a^{(t)}$ denote the video and action latents at denoising step $t$, respectively. After step $N$, the video latent is frozen, and the video-text branch is executed once to build per-layer cached keys and values for the fixed visual-language context. Subsequent denoising steps update only the action tokens: action queries attend to the cached video/text keys and values together with their own action keys and values. This removes the repeated video-stream computation from the latter part of sampling while preserving the same attention semantics as the V2A-style model. Combined with the other inference optimizations above, this design enables \\ours to reach a 11 Hz inference frequency, exceeding typical human reaction speed.

\paragraph{Action Smoothing.} Following DreamZero \cite{ye2026world}, Motubrain applies action chunk smoothing to improve execution smoothness. Each chunk is upsampled to twice its original temporal resolution, smoothed with a Savitzky-Golay filter, and then downsampled back to the original resolution.

\paragraph{Frequency-aware interpolation.}
After smoothing, the action sequence is interpolated according to the ratio between the model action frequency and the low-level control frequency. This frequency-aware interpolation preserves the temporal scale of the model prediction: rapid transitions remain fast during execution, while slower motion segments are expanded over a longer control duration. For manipulation tasks with different motion phases, this helps preserve the velocity profile predicted by the model.

\begin{table}[ht]
\centering
\small
\label{tab:wam_speedup}
\caption{Cumulative speedup from inference optimizations. Each row applies the listed technique on top of all preceding ones. Latency is measured end-to-end on the non-autoregressive model. Per-step latency is reported where applicable.}
\renewcommand{\arraystretch}{1.2}
\begin{tabular}{p{3.2cm} c c c c c}
\toprule
\textbf{Technique} & \textbf{Steps} & \textbf{Per-step (ms)} & \textbf{Latency (s)} & \textbf{Frequency (Hz)} & \textbf{Speedup} \\
\midrule
Baseline                              & 50 & 95.0 & 4.90 & 0.20      & 1.00$\times$ \\
+ Noise sampling          & 30 & 95.0 & 2.90 & 0.34      & 1.69$\times$ \\
+ \texttt{torch.compile} & 30 & 32.7 & 0.98 & 1.02      & 5.00$\times$ \\
+ FP8 quantization                    & 30 & 29.3 & 0.88 & 1.14      & 5.57$\times$ \\
+ DiT cache                   & 30 & --   & 0.20 & 5.00      & 24.5$\times$ \\
+ V2A-style              & 30(action-only) & -- & 0.09 & 11.11 & 54.4$\times$ \\
\bottomrule
\end{tabular}
\vspace{0.5em}
\end{table}

\subsubsection{Real-time Inference and Execution}

Real-time control is essential for robotic deployment, while the inference latency of world action models is usually non-negligible. Motubrain therefore decouples the model inference loop from the robot action execution loop: the controller executes the current action chunk at the target control frequency, while the world action model asynchronously generates the next chunk from the latest observation.

However, directly switching to the newly generated chunk may introduce chunk-boundary discontinuities, such as action regression, velocity jumps, and high-frequency jitter, since adjacent chunks may be generated from different observations and action modes. To reduce such boundary mismatch, we adopt an RTC-inspired strategy \cite{black2025real}. The unexecuted part of the current chunk is used as a constraint for the next generation and fused before denoising, where the latency-affected prefix is treated as a frozen region and the remaining overlapping actions are used as soft constraints.

Let the current action chunk be
\begin{equation}
A^{old}=\{a^{old}_0,a^{old}_1,\ldots,a^{old}_{H-1}\},
\end{equation}
where $H$ denotes the prediction horizon. If a new inference request is launched after $s$ actions have been executed, the remaining actions in the current chunk are
\begin{equation}
A^{remain}=\{a^{old}_s,a^{old}_{s+1},\ldots,a^{old}_{H-1}\}.
\end{equation}

Given the end-to-end inference latency $\delta$ and the controller period $\Delta t$, the inference delay measured in action steps is defined as
\begin{equation}
d=\left\lceil \frac{\delta}{\Delta t} \right\rceil .
\end{equation}
The first $d$ steps of the next chunk correspond to the period in which the previous chunk is still being executed while the new inference is running. These actions are therefore fully constrained by the remaining actions from the previous chunk. For the following overlapping region, we apply a smooth decay weight to gradually reduce the influence of the previous chunk and allow the new prediction to take over.

Let $\rho_i$ denote the normalized progress within the fusion window, where $i$ is the action index in the new chunk, and $L$ denotes the end of the fusion window. The exponential decay function used in our implementation is
\begin{equation}
g(\rho_i)=
\frac{\rho_i\left(e^{\rho_i}-1\right)}{e-1}.
\end{equation}
Since $g(\rho_i)$ increases from 0 to 1, the fusion weight is defined as
\begin{equation}
w_i =
\begin{cases}
1, & 0 \le i < d, \\
1-g(\rho_i), & d \le i < L, \\
0, & i \ge L.
\end{cases}
\end{equation}

The fused action is computed as
\begin{equation}
\tilde{a}^{new}_i
=
w_i a^{remain}_i + (1-w_i)a^{new}_i .
\end{equation}

To improve robustness under variable inference and communication latency, Motubrain maintains a delay queue $Q$ that stores recent inference delays, and the system uses $\hat{d}_{t+1}=\max(Q)$ as a conservative estimate for the next inference request.
This estimated delay determines the length of the frozen prefix. The fusion window is also adjusted accordingly: when the estimated delay increases, more steps are treated as fully constrained; when the estimated delay decreases, more future actions are left to be updated by the new prediction. This makes the asynchronous execution more stable under fluctuating network and model latency.

\section{Evaluations}

In this section, we evaluate \ours in both simulation and real world environment.

\subsection{Evaluation on Simulation Environment}
\label{sec:sim_eval}

Following the protocol of RoboTwin 2.0~\cite{chen2025robotwin}, we adopt a multi-task training setup where all models are trained with 2{,}500 demonstrations collected from clean scenes (50 per task) together with 25{,}000 demonstrations from heavily randomized scenes (500 per task). For training efficiency and temporal consistency across tasks, we downsample videos to 5~Hz and action sequences to 10~Hz. We further conduct ablation studies over multiple policy architectures to analyze how different design choices affect the performance of \ours under the same data and evaluation setting.

\begin{table}[htbp]
\caption{Robotwin 2.0 Results. Following previous works, Motubrain is fine-tuned from pre-trained weights on the official RoboTwin 2.0 dataset (clean + randomized), yielding the evaluation results presented in the table. Motubrain-Non-AR represents non-autoregressive mode.}
\centering
\begin{tabular}{lcc}
\toprule
Model & Clean & Randomized \\
\midrule
\multicolumn{3}{l}{\textbf{\# VLA Based}} \\
$\pi_0$ & 65.9 & 58.4 \\
X-VLA & 72.9 & 72.8 \\
$\pi_{0.5}$ & 82.7 & 76.8 \\
starVLA & 88.2 & 88.3 \\
ABot-M0 & 81.2 & 80.4 \\
LingBot-VLA & 86.5 & 85.3 \\
\midrule
\multicolumn{3}{l}{\textbf{\# World Model Based}} \\
JEPA-VLA & 73.5 & - \\
Motus & 88.7 & 87.0 \\
LingBot-VA & 92.9 & 91.5 \\
Fast-WAM & 91.9 & 91.8 \\
Being-H0.7 & 90.2 & 89.6 \\
\textbf{Motubrain-Non-AR w/o Pretrain, HBridge} & 89.1 & 88.8 \\
\textbf{Motubrain-Non-AR w/o Pretrain} & 89.6 & 89.5 \\
\textbf{Motubrain w/o Pretrain} & 91.5 & 91.3 \\
\textbf{Motubrain-Non-AR} & 91.9 & 92.3 \\
\textbf{Motubrain} & \textbf{95.8} & \textbf{96.1} \\
\bottomrule
\end{tabular}
\label{tab:robotwin-overall}
\end{table}

\begin{table}[htbp]
\caption{Per-task success rates on RoboTwin under clean and randomized evaluation settings.}
\centering
\footnotesize
\setlength{\tabcolsep}{3.5pt} 
\begin{tabular}{l cc cc cc cc cc cc}
\toprule
\multirow{2}{*}{Task} & \multicolumn{2}{c}{$\pi_{0.5}$} & \multicolumn{2}{c}{X-VLA} & \multicolumn{2}{c}{Motus} & \multicolumn{2}{c}{LingBot-VA} & \multicolumn{2}{c}{Fast-WAM} & \multicolumn{2}{c}{Motubrain} \\
\cmidrule(lr){2-3} \cmidrule(lr){4-5} \cmidrule(lr){6-7} \cmidrule(lr){8-9} \cmidrule(lr){10-11} \cmidrule(lr){12-13}
& Clean & Rand. & Clean & Rand. & Clean & Rand. & Clean & Rand. & Clean & Rand. & Clean & Rand. \\
\midrule
Adjust Bottle             & \textbf{100} & 99  & \textbf{100} & 99  & 89  & 93  & 90  & 94  & \textbf{100} & \textbf{100} & 99  & \textbf{100} \\
Beat Block Hammer         & 96  & 93  & 92  & 88  & 95  & 88  & 96  & 98  & 99  & 97  & \textbf{100} & \textbf{100} \\
Blocks Ranking RGB        & 92  & 85  & 83  & 83  & 99  & 97  & 99  & 98  & \textbf{100} & \textbf{100} & \textbf{100} & \textbf{100} \\
Blocks Ranking Size       & 49  & 26  & 67  & 74  & 75  & 63  & 94  & 96  & 94  & 98  & \textbf{100} & \textbf{100} \\
Click Alarmclock          & 98  & 89  & 99  & 99  & \textbf{100} & \textbf{100} & 99  & \textbf{100} & \textbf{100} & \textbf{100} & \textbf{100} & \textbf{100} \\
Click Bell                & 99  & 66  & \textbf{100} & \textbf{100} & \textbf{100} & \textbf{100} & \textbf{100} & \textbf{100} & \textbf{100} & \textbf{100} & \textbf{100} & \textbf{100} \\
Dump Bin Bigbin           & 92  & 97  & 79  & 77  & 95  & 91  & 89  & 96  & 97  & 96  & \textbf{99}  & \textbf{100} \\
Grab Roller               & \textbf{100} & \textbf{100} & \textbf{100} & \textbf{100} & \textbf{100} & \textbf{100} & \textbf{100} & \textbf{100} & \textbf{100} & \textbf{100} & \textbf{100} & \textbf{100} \\
Handover Block            & 66  & 57  & 73  & 37  & 86  & 73  & 99  & 78  & 95  & 81  & \textbf{100} & \textbf{95}  \\
Handover Mic              & 98  & 97  & 0   & 0   & 78  & 63  & 94  & 96  & 99  & \textbf{100} & \textbf{100} & \textbf{100} \\
Hanging Mug               & 18  & 17  & 23  & 27  & 38  & 38  & 40  & 28  & \textbf{58}  & \textbf{62}  & 55  & 43  \\
Lift Pot                  & 96  & 85  & 99  & \textbf{100} & 96  & 99  & \textbf{100} & 99  & \textbf{100} & \textbf{100} & \textbf{100} & \textbf{100} \\
Move Can Pot              & 51  & 55  & 89  & 86  & 34  & 74  & 94  & 97  & 90  & 88  & \textbf{99}  & \textbf{100} \\
Move Pillbottle Pad       & 84  & 61  & 73  & 71  & 93  & 96  & 99  & 99  & \textbf{100} & 99  & \textbf{100} & \textbf{100} \\
Move Playingcard Away     & 96  & 84  & 93  & 98  & \textbf{100} & 96  & \textbf{100} & 99  & \textbf{100} & \textbf{100} & \textbf{100} & \textbf{100} \\
Move Stapler Pad          & 56  & 42  & 78  & 73  & 83  & 85  & \textbf{91}  & 79  & 77  & 64  & 85  & \textbf{93}  \\
Open Laptop               & 90  & 96  & 93  & \textbf{100} & 95  & 91  & 92  & 94  & 98  & \textbf{100} & \textbf{100} & 99  \\
Open Microwave            & 34  & 77  & 79  & 71  & 95  & 91  & 82  & 86  & 62  & 45  & \textbf{100} & \textbf{100} \\
Pick Diverse Bottles      & 81  & 71  & 58  & 36  & 90  & \textbf{91}  & 89  & 82  & 80  & 85  & \textbf{95}  & 89  \\
Pick Dual Bottles         & 93  & 63  & 47  & 36  & 96  & 90  & \textbf{100} & 99  & \textbf{100} & 96  & \textbf{100} & \textbf{100} \\
Place A2B Left            & 87  & 82  & 48  & 49  & 88  & 79  & 97  & 93  & 95  & 93  & \textbf{100} & \textbf{99}  \\
Place A2B Right           & 87  & 84  & 36  & 36  & 91  & 87  & \textbf{97}  & 95  & 93  & \textbf{99}  & 95  & \textbf{99}  \\
Place Bread Basket        & 77  & 64  & 81  & 71  & 91  & 94  & \textbf{97}  & 95  & 91  & 93  & \textbf{97}  & \textbf{99}  \\
Place Bread Skillet       & 85  & 66  & 77  & 67  & 86  & 83  & \textbf{95}  & 90  & 90  & \textbf{93}  & \textbf{95}  & \textbf{93}  \\
Place Burger Fries        & 94  & 87  & 94  & 94  & 98  & 98  & 97  & 95  & 96  & 99  & \textbf{99}  & \textbf{100} \\
Place Can Basket          & 62  & 62  & 49  & 52  & 81  & 76  & 81  & 84  & 71  & 69  & \textbf{85}  & \textbf{93}  \\
Place Cans Plasticbox     & 94  & 84  & 97  & 98  & 98  & 94  & \textbf{100} & 99  & 99  & 96  & \textbf{100} & \textbf{100} \\
Place Container Plate     & \textbf{99}  & 95  & 97  & 95  & 98  & 99  & \textbf{99}  & 97  & 96  & \textbf{100} & 97  & \textbf{100} \\
Place Dual Shoes          & 75  & 75  & 79  & 88  & 93  & 87  & \textbf{94}  & 89  & \textbf{94}  & 88  & 93  & \textbf{95}  \\
Place Empty Cup           & \textbf{100} & 99  & \textbf{100} & 98  & 99  & 98  & \textbf{100} & \textbf{100} & \textbf{100} & \textbf{100} & \textbf{100} & \textbf{100} \\
Place Fan                 & 87  & 85  & 80  & 75  & 91  & 87  & \textbf{99}  & 93  & 96  & 96  & 97  & \textbf{97}  \\
Place Mouse Pad           & 60  & 39  & 70  & 70  & 66  & 68  & 93  & 96  & 83  & 89  & \textbf{97}  & \textbf{97}  \\
Place Object Basket       & 80  & 76  & 44  & 39  & 81  & 87  & \textbf{91}  & 88  & 89  & 88  & 89  & \textbf{93}  \\
Place Object Scale        & 86  & 80  & 52  & 74  & 88  & 85  & 96  & 95  & 90  & 97  & \textbf{97}  & \textbf{100} \\
Place Object Stand        & 91  & 85  & 86  & 88  & 98  & \textbf{97}  & \textbf{99}  & 96  & 90  & 94  & 97  & \textbf{97}  \\
Place Phone Stand         & 81  & 81  & 88  & 87  & 87  & 86  & \textbf{97}  & 97  & \textbf{97}  & \textbf{99}  & \textbf{97}  & 93  \\
Place Shoe                & 92  & 93  & 96  & 95  & 99  & 97  & 98  & 98  & 96  & \textbf{99}  & \textbf{100} & 97  \\
Press Stapler             & 87  & 83  & 92  & 98  & 93  & 98  & 85  & 82  & 90  & 97  & \textbf{100} & \textbf{100} \\
Put Bottles Dustbin       & 84  & 79  & 74  & 77  & 81  & 79  & 87  & \textbf{91}  & \textbf{95}  & 90  & 87  & 87  \\
Put Object Cabinet        & 80  & 79  & 46  & 48  & 88  & 71  & 85  & 87  & \textbf{94}  & 89  & 92  & \textbf{93}  \\
Rotate QRcode             & 89  & 87  & 34  & 33  & 89  & 73  & \textbf{96}  & 91  & 93  & 89  & 90  & \textbf{93}  \\
Scan Object               & 72  & 65  & 14  & 36  & 67  & 66  & \textbf{96}  & 91  & 89  & \textbf{92}  & 94  & 90  \\
Shake Bottle              & 99  & 97  & \textbf{100} & \textbf{100} & \textbf{100} & 97  & \textbf{100} & 97  & \textbf{100} & \textbf{100} & \textbf{100} & \textbf{100} \\
Shake Bottle Horizontally & 99  & 99  & 99  & \textbf{100} & \textbf{100} & 98  & \textbf{100} & 99  & \textbf{100} & \textbf{100} & \textbf{100} & \textbf{100} \\
Stack Blocks Three        & 91  & 76  & 6   & 10  & 91  & 95  & 99  & 98  & 95  & 97  & \textbf{100} & \textbf{100} \\
Stack Blocks Two          & 97  & \textbf{100} & 92  & 87  & \textbf{100} & 98  & \textbf{100} & 98  & \textbf{100} & \textbf{100} & \textbf{100} & \textbf{100} \\
Stack Bowls Three         & 77  & 71  & 76  & 86  & 79  & 87  & \textbf{86}  & 83  & 80  & 81  & \textbf{86}  & \textbf{90}  \\
Stack Bowls Two           & 95  & 96  & 96  & 93  & \textbf{98}  & \textbf{98}  & 94  & \textbf{98}  & 92  & \textbf{98}  & 92  & \textbf{98}  \\
Stamp Seal                & 79  & 55  & 76  & 82  & 93  & 92  & 96  & 97  & 90  & 94  & \textbf{100} & \textbf{98}  \\
Turn Switch               & 62  & 54  & 40  & 61  & \textbf{84}  & 78  & 44  & 45  & 61  & 59  & 82  & \textbf{84}  \\
\midrule
Average                   & 82.74 & 76.76 & 72.80 & 72.84 & 88.66 & 87.02 & 92.90 & 91.50 & 91.88 & 91.78 & \textbf{95.80} & \textbf{96.08} \\
\bottomrule
\end{tabular}
\label{tab:robotwin-per-task}
\end{table}

As shown in Table~\ref{tab:robotwin-overall}, \ours achieves the best average success rate on RoboTwin, reaching 95.8\% in clean scenes and 96.1\% in randomized scenes. It ranks first in both settings and is the only model on the leaderboard whose average score exceeds 95\% under randomized evaluation. At the per-task level, as summarized in Table~\ref{tab:robotwin-per-task}, Motubrain attains a perfect score on 24 tasks under the clean setting and 25 tasks under the randomized setting, with 19 tasks achieving 100\% success in both settings. Moreover, it surpasses 90\% success on 42 clean tasks and 44 randomized tasks, demonstrating consistently strong generalization across the full 50-task benchmark and strong ability to act robustly in the world.

From the detailed breakdown in Table~\ref{tab:robotwin-per-task}, the largest gains are concentrated on tasks that require robust multi-stage manipulation and precise interaction under visual variation. In particular, we observe clear improvements on handover and coordination-heavy tasks such as \textit{Handover Block}, articulated-object interaction tasks such as \textit{Open Microwave}, \textit{Press Stapler}, and \textit{Turn Switch}, as well as fine-grained spatial arrangement tasks including \textit{Blocks Ranking Size}, \textit{Move Can Pot}, \textit{Place A2B Left}, and \textit{Place Can Basket}. These results suggest that \ours is especially effective on tasks demanding accurate temporal modeling, stable contact-rich control, and robustness to scene randomization.

\begin{figure*}[htbp]
\centering
\begin{minipage}[t]{0.48\textwidth}
  \centering
  \includegraphics[width=\linewidth]{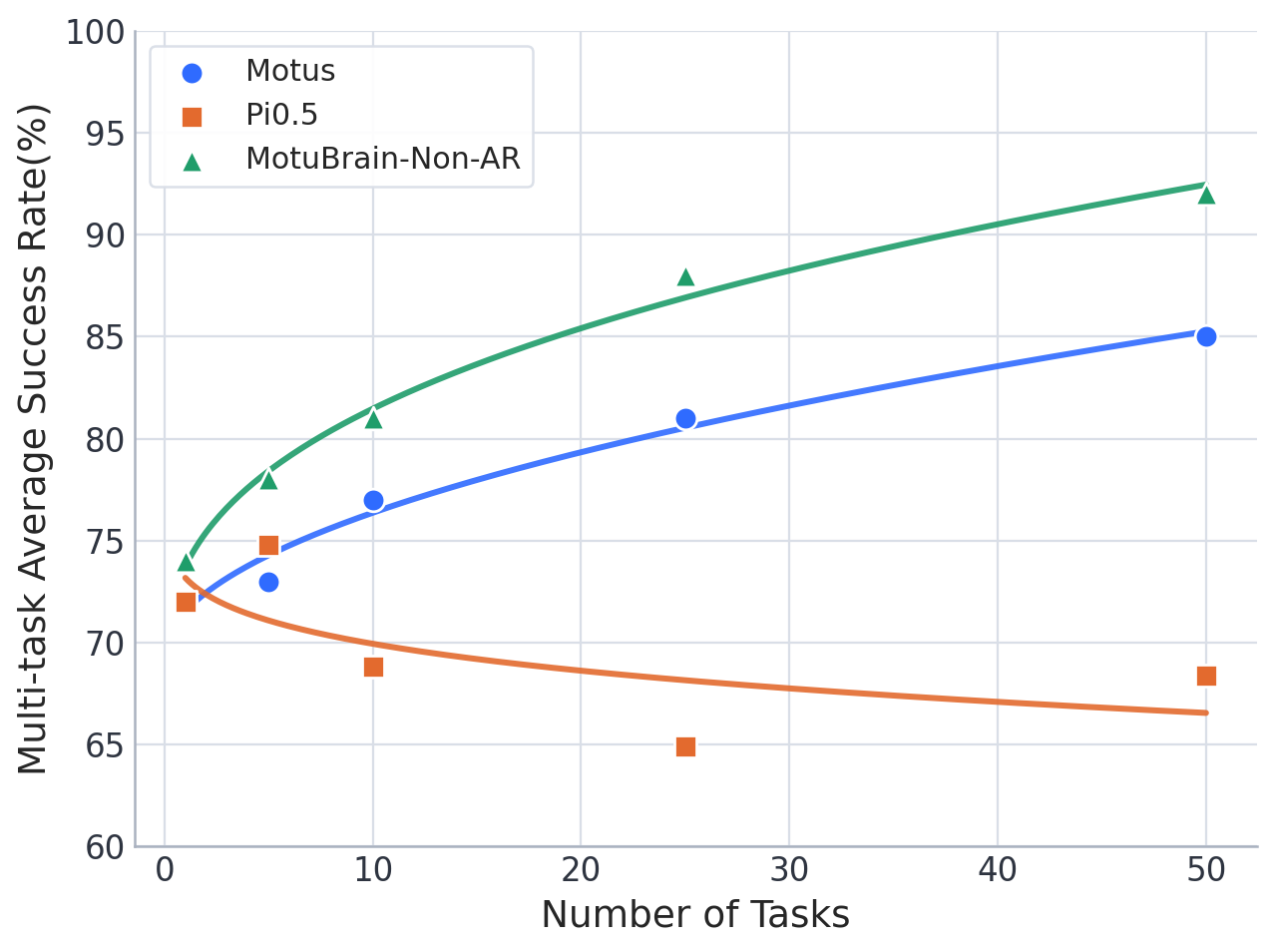}
  \captionof{figure}{Task scaling results. For each point on the curve, we train the model using data from only the specified number of tasks and keep optimization running until convergence, i.e., until the average success rate becomes approximately stable. As the number of training tasks increases, \ours shows a clear upward trend in average success rate and consistently stronger scaling behavior than conventional VLA baselines.}
  \label{fig:task-scaling}
\end{minipage}
\hfill
\begin{minipage}[t]{0.48\textwidth}
  \centering
  \includegraphics[width=\linewidth]{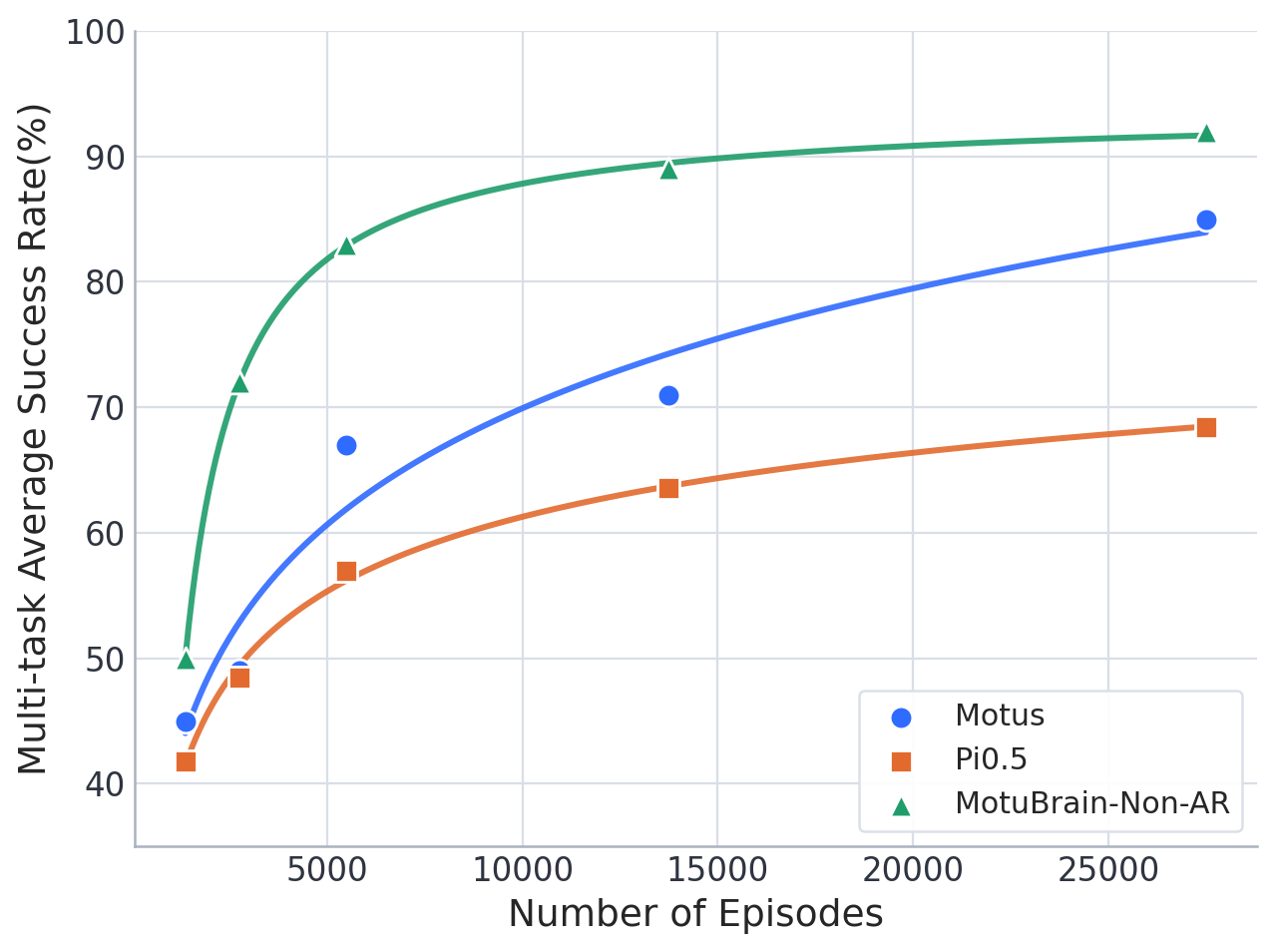}
  \captionof{figure}{Data scaling results. For each data budget, we uniformly subsample demonstrations from every task to keep the per-task data distribution balanced. The number of training steps is set proportional to the total amount of training data; in particular, training on the full 27{,}500-trajectory dataset uses 50{,}000 optimization steps. \ours continues to benefit from additional training data, indicating that larger-scale supervision improves policy performance and robustness.}
  \label{fig:data-scaling}
\end{minipage}
\end{figure*}

We further observe a favorable multi-task scaling trend: as the number of training tasks increases, the average success rate of \ours continues to improve, suggesting that more tasks expose the model to more reusable world knowledge about objects, contacts, and temporal transitions, as illustrated in Figure~\ref{fig:task-scaling}. In contrast, conventional VLA baselines exhibit noticeably weaker scaling in the same regime, indicating stronger task interference and less effective knowledge sharing across tasks. We also find that \ours is substantially more data-efficient than both conventional VLA baselines and the previous Motus model: under the same data scaling protocol, Motubrain reaches stronger performance with fewer training trajectories and continues to improve more effectively as data grows, as shown in Figure~\ref{fig:data-scaling}.

Taken together, Figures~\ref{fig:task-scaling} and~\ref{fig:data-scaling} further suggest that, in our setting, increasing task diversity is more effective than merely scaling up the amount of data collected for a fixed task set, as evidenced by the steeper improvement trend in the task-scaling curve. This observation is consistent with the hypothesis that broader task coverage exposes the model to a richer set of interaction patterns, object affordances, and temporal transitions, thereby improving knowledge reuse and cross-task generalization more efficiently than data duplication alone.

\subsection{Evaluation on World Models}
\label{sec:worldarena}

Beyond action execution, we also evaluate whether \ours can understand and predict changes in the world itself. We adopt WorldArena~\cite{shang2026worldarena}, a unified benchmark that scores embodied world models through 16 numerical metrics across six perceptual sub-dimensions---visual quality, motion quality, content consistency, physics adherence, 3D accuracy, and controllability. We use the entire 2{,}500-trajectory release for high-resolution training and report benchmark submission results. Because \ours is a unified world-action model, the same backbone simultaneously serves as the manipulation policy used in Section~\ref{sec:sim_eval} and as the world predictor evaluated here. For WorldArena, we run the model in its forward-dynamics mode (FDM), which serves as an action-conditioned world model. We train and run inference at high resolution with a $5$\,Hz video stream paired with $10$\,Hz actions ($1{:}2$ ratio), using classifier-free guidance to improve instruction-following capability.

\begin{figure}[H]
  \centering
  \includegraphics[width=\linewidth]{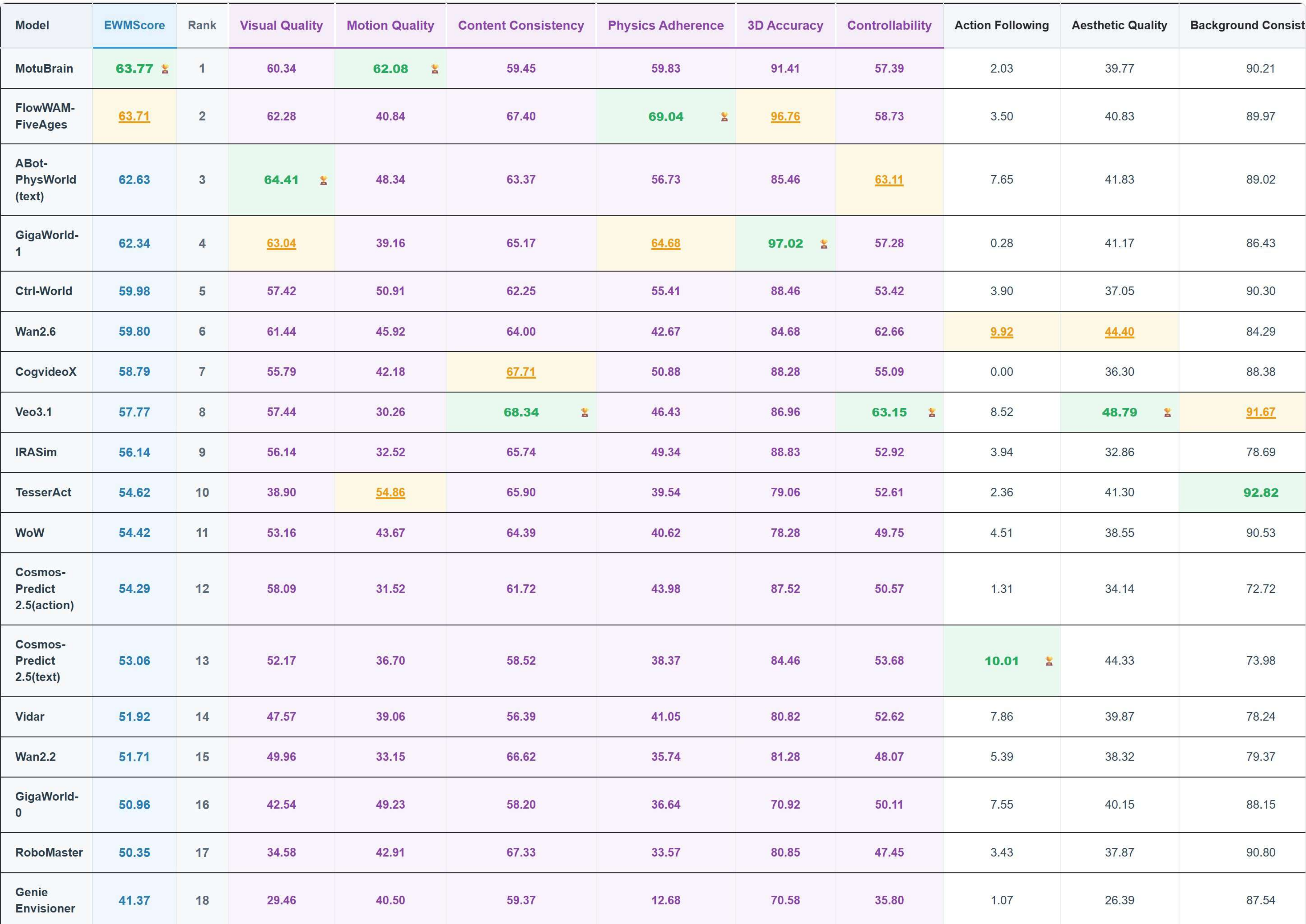}
  \caption{WorldArena public leaderboard.}
  \label{fig:worldarena_leaderboard}
\end{figure}

As shown in Figure~\ref{fig:worldarena_leaderboard}, \ours attains the highest EWMScore (\textbf{63.77}) among embodied world models and video baselines on the WorldArena leaderboard. Its margin over the strongest video-generation baseline shown here (Wan2.6, $59.80$) is approximately four points. Notably, \ours ranks first on motion quality and remains competitive on several other dimensions, indicating that strong dynamics modeling can be achieved without sacrificing overall perceptual quality.

The detailed per-metric breakdown in Table~\ref{tab:wa-overall} shows that \ours's lead is driven primarily by its strong performance in the \textbf{Motion Quality} dimension. The Flow Score, which measures the overall magnitude of motion in the predicted video, is the highest among all baselines, indicating that \ours generates sequences with sustained inter-frame displacement rather than near-static clips that collapse to a stable but trivial solution. The leading Motion Smoothness indicates that the produced motion unfolds with realistic inertia rather than frame-to-frame jitter. The high Dynamic Degree, which measures whether motion concentrates in the regions that actually matter, further confirms that \ours's predicted activity is localized on the embodiment and the manipulated objects rather than diffused into background flicker. Together, these three metrics show that \ours predicts motion that is substantial, smooth, and concentrated in embodiment-relevant regions.


This distinction matters because the original WorldArena study reports that EWMScore correlates only weakly with downstream action-planning success ($r{=}0.36$), reflecting the well-known perception--functionality gap: visually impressive world models often fail when used for control, whereas functionally useful ones often appear visually unpolished. \ours helps bridge this gap. On the perceptual side, it ranks first on the EWMScore leaderboard among embodied world models and video baselines; on the functional side, it achieves an average success rate of $95.8\%$ across the 50 RoboTwin manipulation tasks (Section~\ref{sec:sim_eval}), the highest among all evaluated VLA and world-model baselines. This cross-benchmark consistency suggests that the dynamics representation \ours learns is simultaneously perceptually faithful and functionally actionable when consumed by a policy.

\begin{table*}[htbp]
\caption{Per-metric WorldArena results and overall EWMScore comparison. We compare \ours with several publicly listed and representative entries on the WorldArena leaderboard. The best score in each row is shown in bold. EWMScore is the arithmetic mean of all 16 normalized metrics, scaled to $[0,100]$.}
\centering
\footnotesize
\setlength{\tabcolsep}{4pt}
\renewcommand{\arraystretch}{1.05}
\begin{tabular}{ll c ccccc}
\toprule
\multicolumn{2}{l}{} & \textbf{\ours} & Veo3.1 & Wan2.6 & Ctrl-World & ABot-PW & GigaWorld-1 \\
\midrule
\multirow{3}{*}{\parbox{1.8cm}{\centering Visual\\Quality}}
& Image Quality            & 0.4459 & 0.6557 & \textbf{0.6736} & 0.4244 & 0.6103 & 0.5118 \\
& Aesthetic Quality        & 0.3977 & \textbf{0.4879} & 0.4440 & 0.3705 & 0.4183 & 0.4117 \\
& JEPA Similarity          & 0.9667 & 0.5797 & 0.7257 & 0.9277 & 0.9036 & \textbf{0.9677} \\
\midrule
\multirow{3}{*}{\parbox{1.8cm}{\centering Motion\\Quality}}
& Dynamic Degree           & \textbf{0.5148} & 0.1466 & 0.3363 & 0.4182 & 0.3933 & 0.3052 \\
& Flow Score               & \textbf{0.4911} & 0.0826 & 0.2201 & 0.3357 & 0.2922 & 0.1864 \\
& Motion Smoothness        & \textbf{0.8566} & 0.6785 & 0.8212 & 0.7734 & 0.7648 & 0.6833 \\
\midrule
\multirow{3}{*}{\parbox{1.8cm}{\centering Content\\Consistency}}
& Subject Consistency      & 0.8240 & 0.7582 & 0.7315 & \textbf{0.8356} & 0.8056 & 0.8097 \\
& Background Consistency   & 0.9021 & \textbf{0.9167} & 0.8429 & 0.9030 & 0.8902 & 0.8643 \\
& Photometric Consistency  & 0.0574 & \textbf{0.3752} & 0.3457 & 0.1288 & 0.2052 & 0.2811 \\
\midrule
\multirow{2}{*}{\parbox{1.8cm}{\centering Physics Adherence}}
& Interaction Quality      & 0.7174 & 0.8150 & 0.7316 & 0.6262 & \textbf{0.8196} & 0.7510 \\
& Trajectory Accuracy      & 0.4793 & 0.1136 & 0.1218 & 0.4820 & 0.3150 & \textbf{0.5427} \\
\midrule
\multirow{2}{*}{\parbox{1.8cm}{\centering 3D Accuracy}}
& Depth Accuracy           & 0.8992 & 0.7427 & 0.7543 & 0.9325 & 0.7199 & \textbf{0.9844} \\
& Perspectivity            & 0.9290 & \textbf{0.9964} & 0.9394 & 0.8366 & 0.9894 & 0.9560 \\
\midrule
\multirow{3}{*}{\parbox{1.8cm}{\centering Controllability}}
& Instruction Following    & 0.8072 & \textbf{0.9714} & 0.8996 & 0.6768 & 0.9210 & 0.8214 \\
& Semantic Alignment       & 0.8941 & 0.8379 & 0.8809 & 0.8868 & \textbf{0.8958} & 0.8942 \\
& Action Following         & 0.0203 & 0.0852 & \textbf{0.0992} & 0.0390 & 0.0765 & 0.0028 \\
\midrule
\multicolumn{2}{l}{\textbf{EWMScore} ($\uparrow$)}
                          & \textbf{63.77} & 57.77 & 59.80 & 59.98 & 62.63 & 62.34 \\
\bottomrule
\end{tabular}
\label{tab:wa-overall}
\end{table*}

\FloatBarrier

\subsection{Real-World Experiments}

We further investigate rapid deployment to new embodiments in real-world settings. Starting from a pretrained model, \ours can be adapted to a new embodiment using only 50–100 same-embodiment trajectories. We validate this capability across multiple humanoid platforms. Notably, these results are achieved without relying on auxiliary components such as VLM-based planners, dual-system decompositions, external memory modules, or additional reinforcement/retry data. In other words, the native world action model alone provides a sufficiently strong control prior to enable practical transfer.

\subsubsection{Real-World Performance}

\paragraph{Quantitative Evaluation.}

We conduct a quantitative evaluation of \ours on a suite of representative real-world household tasks. Performance is measured using a normalized scoring metric with a maximum value of 100, where each sub-task step contributes equally to the overall score. A step receives full credit if it is successfully completed on the first attempt; if it succeeds after one retry, two retries, or three or more retries, it receives 80\%, 50\%, or 0 credit, respectively.

For each task, the model is trained on 100 task-specific trajectories consisting of single, coherent, and successfully executed compound actions, without atomic-level annotations. We further assess performance across tasks with varying execution horizons and differing numbers of atomic actions. The quantitative results are summarized in Table~\ref{tab:realworld-benchmark-pouring}, Table~\ref{tab:realworld-benchmark-cocktail}, and Table~\ref{tab:realworld-benchmark-flower}.

In the Making Oden task, which requires simultaneous bimanual manipulation with one arm pouring a drink and the other scooping dumplings, \ours is evaluated over five consecutive trials, with an average execution time of 33 seconds across seven atomic actions. The robot executes the subtasks concurrently with its left and right hands, achieving an overall score of 98.54, with only a single atomic action failure observed.

\begin{table*}[htbp]
\caption{Per-step performance evaluation of the bimanual Making Oden task. The overall score reaches 98.54, with only a slight deficiency in the pouring step.}
\centering
\footnotesize
\setlength{\tabcolsep}{3.5pt}
\begin{tabular}{p{2cm} c p{6cm} cc}
\toprule
\multicolumn{1}{c}{Task Name} & \multicolumn{1}{c}{IDs} & \multicolumn{1}{c}{Sub-task Names} & \multicolumn{1}{c}{Total Scores} & \multicolumn{1}{c}{Our Scores} \\
\midrule
\multirow{7}{*}{\parbox[c][\dimexpr\baselineskip*7\relax][c]{2.0cm}{\centering\arraybackslash Making Oden}} & 1 & Grab the drink with your right hand & 14.29 & 14.29 \\
 & 2 & Pours drink successfully with your right hand & 14.29 & 12.83 \\
 & 3 & Put the beverage bottle down with your right hand & 14.29 & 14.29 \\
 & 4 & Grab the spoon with your left hand & 14.29 & 14.29 \\
 & 5 & Pick up the balls with your left hand & 14.29 & 14.29 \\
 & 6 & Place the balls in the bowl with your left hand & 14.29 & 14.29 \\
 & 7 & Put the spoon back with your left hand & 14.26 & 14.26 \\
\bottomrule
\end{tabular}

\label{tab:realworld-benchmark-pouring}
\end{table*}

In the Mixing Cocktails task, which involves an extended long-horizon sequence with complex multi-step dependencies across 15 atomic actions, the model is evaluated over an average execution time of 124 seconds. Across seven consecutive trials, \ours attains an overall score of 97.34, demonstrating stable performance under substantially increased temporal complexity.

\begin{table*}[htbp]
\caption{Per-step real-world evaluation results for the long-horizon Mixing Cocktails task. Each sub-task contributes equally to the final score, and \ours achieves strong overall performance with only minor deficiencies in a small number of steps.}
\centering
\footnotesize
\setlength{\tabcolsep}{3.5pt}
\begin{tabular}{p{2cm} c p{7cm} cc}
\toprule
\multicolumn{1}{c}{Task Name} & \multicolumn{1}{c}{IDs} & \multicolumn{1}{c}{Sub-task Names} & \multicolumn{1}{c}{Total Scores} & \multicolumn{1}{c}{Our Scores} \\
\midrule
\multirow{15}{*}{\parbox[c][\dimexpr\baselineskip*15\relax][c]{2.0cm}{\centering Mixing Cocktails}} & 1 & Pick up the bottle & 6.67 & 6.67 \\
 & 2 & Fill the mixing glass & 6.67 & 5.72 \\
 & 3 & Return the bottle to its original position & 6.67 & 6.67 \\
 & 4 & Grasp the milk bottle & 6.67 & 6.67 \\
 & 5 & Pour milk into the glass & 6.67 & 6.29 \\
 & 6 & Return the milk bottle to its original position & 6.67 & 6.67 \\
 & 7 & Grasp the mixing glass & 6.67 & 6.67 \\
 & 8 & Pour the mixed liquid into the serving glass & 6.67 & 6.67 \\
 & 9 & Place the mixing glass back & 6.67 & 6.67 \\
 & 10 & Grasp a peppermint leaf & 6.67 & 6.67 \\
 & 11 & Place the peppermint leaf into the glass & 6.67 & 6.67 \\
 & 12 & Grasp the prepared drink & 6.67 & 6.67 \\
 & 13 & Place the prepared drink on the serving tray & 6.67 & 5.53 \\
 & 14 & Squeeze the syrup bottle to complete the mixing & 6.67 & 6.48 \\
 & 15 & Return to the home position & 6.62 & 6.62 \\
\bottomrule
\end{tabular}
\label{tab:realworld-benchmark-cocktail}
\end{table*}

In the Flower Arrangement task, which requires precise and fine-grained manipulation for accurate flower placement, \ours is evaluated over 10 consecutive trials, achieving an average execution time of 138 seconds across 10 atomic actions and an overall score of 83.30. Notably, for repetitive failure cases, \ours demonstrates an inherent retry capability despite the absence of explicit recovery supervision during training, indicating that it captures the underlying long-horizon task objective rather than relying on fixed motion replay.

\begin{table*}[htbp]
\caption{Per-step real-world evaluation results for the \textit{Flower Arrangement} task. Our model demonstrates stable execution across repeated long-horizon trials.}
\centering
\footnotesize
\setlength{\tabcolsep}{3.5pt}
\begin{tabular}{p{2cm} c p{8cm} cc}
\toprule
\multicolumn{1}{c}{Task Name} & \multicolumn{1}{c}{IDs} & \multicolumn{1}{c}{Sub-task Names} & \multicolumn{1}{c}{Total Scores} & \multicolumn{1}{c}{Our Scores} \\
\midrule
\multirow{10}{*}{\parbox[c][\dimexpr\baselineskip*10\relax][c]{2.0cm}{\centering\arraybackslash Flower Arrangement}} & 1 & Left hand picks up the first flower branch & 10 & 8.8 \\
 & 2 & Insert flowers into vase & 10 & 8.8 \\
 & 3 & Pick up the second branch with your left hand & 10 & 7.3 \\
 & 4 & Insert flowers into vase & 10 & 8.0 \\
 & 5 & Pick up the third flower branch with your left hand & 10 & 5.6 \\
 & 6 & Insert flowers into vase & 10 & 5.8 \\
 & 7 & Pick up the kettle with your right hand and aim it at the vase & 10 & 10 \\
 & 8 & Hold water spray with left hand & 10 & 9.5 \\
 & 9 & Release the kettle with your left hand and lower the kettle with your right hand & 10 & 10 \\
 & 10 & Left hand delivering the vase to the table & 10 & 9.5 \\
\bottomrule                                                                                       
\end{tabular}
\label{tab:realworld-benchmark-flower}
\end{table*}

\paragraph{Qualitative Analysis.}

We further evaluate \ours in real-world home environments on several everyday tasks, including flower arrangement, meal preparation, cocktail mixing, sofa tidying, and sink organization. These tasks encompass representative activities across multiple household environments. Across all settings, \ours consistently accomplishes them on diverse humanoid robot embodiments, demonstrating robust generalization and the ability to reliably compose and execute fundamental multi-step household skills.

\textbf{Few-shot Evaluation and Self-correction.}
    Following the training-stage setup, we utilize only 100 trajectories of continuous atomic actions, without any additional sub-task annotations. In the flower-arrangement task, where the placement height varies across trials, \ours is able to execute repeated long-horizon task sequences in a novel home environment, with each rollout lasting approximately 2-3 minutes. In these experiments, the model receives the goal instruction only once at the beginning of each episode. During execution, it predicts future action states and continuously refines its policy through closed-loop integration of visual observations. When execution errors occur, such as failed insertion of a flower into the vase, the model leverages updated perceptual feedback to adapt its behavior and perform online correction. This long-horizon, few-shot evaluation demonstrates that \ours possesses strong physical world modeling capabilities and achieves a significantly higher success rate than Vision-Language-Action models.

\textbf{Bimanual Action Generalization.}
    Our demonstrations include asymmetric bimanual behaviors, where the two arms execute distinct subtasks concurrently, such as pouring and scooping. Our model is robust to asynchronous subtask completion: when one subtask is externally fulfilled ahead of time, the other arm continues execution without interruption. For example, if a person helps fill the drink before the robot finishes pouring, the other arm can still continue the scooping behavior and complete the remaining sub-task. Such behavior suggests that \ours does not rely on rigid inter-arm dependencies, but instead learns coordinated yet decoupled control under a shared temporal context.

\textbf{Scene Generalization.}
    We evaluate cross-scene generalization under a limited-data setting. In the flower-arrangement task, our model trained on a single scene is deployed in a novel environment with previously unseen flowers and vases. Despite being trained on only one type of flower and vase, the model generalizes to four unseen flower-vase combinations and achieves a success rate above 80\%. In contrast, the VLA-based baseline typically requires training on at least three object categories with diverse shapes and sizes before achieving reasonable generalization to a fourth unseen instance. These results indicate that \ours captures object geometry of novel household objects and leverages it for downstream action reasoning.

\subsubsection{Real-World Demonstrations}

We visualize the real-world deployment of \ours across a range of manipulation tasks. As shown in Figure \ref{fig:real-world-demos}, the model executes these tasks reliably and effectively in diverse household scenarios.

\begin{center}
\noindent\begin{minipage}{\linewidth}
  \centering
  \includegraphics[width=0.92\linewidth]{src/figs/b_Bathroom_1.jpg}

  {\small Place the toothbrush in the cup and put the soap back in its place.}
\end{minipage}

\par\vspace{8pt}

\noindent\begin{minipage}{\linewidth}
  \centering
  \includegraphics[width=0.92\linewidth]{src/figs/b_Cocktail_vstack.jpg}

  {\small Mix a cocktail using the milk and beverage on the table, place it on the tray, and serve it to the customer.}
\end{minipage}

\par\vspace{8pt}

\noindent\begin{minipage}{\linewidth}
  \centering
  \includegraphics[width=0.92\linewidth]{src/figs/b_Oden_1.jpg}

  {\small Pour a glass of juice while scooping a serving of dumplings from the pot into the bowl.}
\end{minipage}

\par\vspace{8pt}

\noindent\begin{minipage}{\linewidth}
  \centering
  \includegraphics[width=0.92\linewidth]{src/figs/b_Sofa_vstack.jpg}

  {\small Put the clothes on the sofa into the laundry basket and put the pillows back in place.}
\end{minipage}

\par\vspace{8pt}

\noindent\begin{minipage}{\linewidth}
  \centering
  \includegraphics[width=0.92\linewidth]{src/figs/b_flower_vstack.jpg}

  {\small Insert the flowers into the vase, then spray water with the watering can.}
\end{minipage}

\captionof{figure}{Real-world demonstrations on multiple long-horizon and dexterous manipulation tasks.}
\label{fig:real-world-demos}
\end{center}





\section{Conclusion and Future Work}

We present \ours, a unified world action model that learns robot control by jointly modeling future visual dynamics and action generation. By combining large-scale pretraining with lightweight robot adaptation, \ours inherits broad semantic and physical priors while remaining practical for downstream deployment. Empirically, the model achieves strong and consistent performance across three complementary settings: large-scale simulation benchmarks, open-ended world-model evaluation, and real-world humanoid transfer. Together, these results suggest that a single pretrained world-action model can serve as a scalable foundation for both understanding how the world evolves and deciding how a robot should act within it.

Our study also highlights a broader design principle for embodied intelligence: action learning benefits substantially from being trained together with predictive world modeling rather than being optimized as an isolated imitation problem. The resulting representation is not only more robust to compounding execution errors, but also more transferable across tasks, embodiments, and environments.

There remain several important directions for future work. First, current adaptation still relies on a modest amount of same-embodiment robot data; reducing this requirement further would improve accessibility and deployment speed. Second, extending the framework to longer-horizon mobile manipulation, richer tactile interaction, and more dynamic human-centered environments would better test the limits of the learned world prior. Finally, integrating stronger uncertainty estimation, explicit memory, and online test-time adaptation may further improve robustness in open-world settings where disturbances and task variations cannot be fully covered during training.

\enlargethispage{3\baselineskip}
\section{Contributors}
\small








\textbf{Data:} Chendong Xiang*, Louis Liu*, Jiabao Liu*, James Li*, Zeyuan Wang, Hongzhe Bi, Hengkai Tan 

\textbf{Base Model:} Zeyuan Wang*, Chendong Xiang*, Hengkai Tan*, Haitian Liu, Yao Feng, Ruowen Zhao, Shuhe Huang, Hongzhe Bi

\textbf{Post-Training:} Zeyuan Wang*, Chendong Xiang*, Haitian Liu*, Rongxu Cui*, Ruowen Zhao, Hengkai Tan, Jingrui Pang, Yao Feng, Shuhe Huang, Mengchen Cai, Yinze Rong

\textbf{Evaluation:} Rongxu Cui*, Zeyuan Wang*, Haitian Liu*, Chendong Xiang*, Hengkai Tan*, Mengchen Cai*, Ruowen Zhao*, Shuhe Huang*, Runqing Wang, Kiro Jing, James Li, Yao Feng, Yinze Rong

\textbf{Project Lead:} Hengkai Tan

\textbf{Advisor:} Fan Bao, Jun Zhu

\textit{* denotes the core-contributors or leaders of each sub-module.}
\normalsize

\bibliographystyle{plain}
\bibliography{main}

\end{document}